\documentclass[runningheads]{llncs}

\usepackage[T1]{fontenc}
\usepackage{graphicx}%
\usepackage{multirow}%
\usepackage{amsmath,amssymb,amsfonts}%
\usepackage{mathrsfs}%
\usepackage[title]{appendix}%
\usepackage{xcolor}%
\usepackage{textcomp}%
\usepackage{manyfoot}%
\usepackage{booktabs}%
\usepackage{algorithm}%
\usepackage{algorithmicx}%
\usepackage{algpseudocode}%
\usepackage{listings}%
\usepackage{array}

\definecolor{SoftRed}{RGB}{192, 44, 56}    
\definecolor{SoftGreen}{RGB}{34, 139, 34}  

\newcommand{\plus}[1]{\textcolor{SoftGreen}{+}#1}
\newcommand{\minus}[1]{\textcolor{SoftRed}{- }#1}
\definecolor{DimBlue}{RGB}{70, 130, 180}    
\definecolor{DimOrange}{RGB}{210, 105, 30}  
\definecolor{DimGreen}{RGB}{85, 107, 47}    
\definecolor{DimPurple}{RGB}{147, 112, 219} 
\definecolor{DimRed}{RGB}{178, 34, 34}      
\definecolor{DimTeal}{RGB}{0, 128, 128}     

\begin{document}

\title{SLOW: Strategic Logical-inference Open Workspace for Cognitive Adaptation in AI Tutoring}

\author{Yuang Wei\inst{1}\orcidID{0000-0002-8187-4011}\thanks{Equal contribution(\email{philrain.cs@gmail.com, rj.yuzu.li@gmail.com}).} \and
Ruijia Li\inst{1}\orcidID{0009-0001-9680-7508}\protect\footnotemark[1] \and
Bo Jiang\inst{1}\orcidID{0000-0002-7914-1978}\thanks{Corresponding author(\email{bjiang@deit.ecnu.edu.cn}).}}
\authorrunning{Y. Wei et al.}
\titlerunning{SLOW: Strategic Logical-inference Open Workspace}
\institute{Shanghai Institute of Artificial Intelligence for Education, Shanghai, China}

\maketitle
\begin{abstract}
While Large Language Models (LLMs) have demonstrated remarkable fluency in educational dialogues, most generative tutors primarily operate through intuitive, single-pass generation. This reliance on fast thinking precludes a dedicated reasoning workspace, forcing multiple diagnostic and strategic signals to be processed in a conflated manner. As a result, learner cognitive diagnosis, affective perception, and pedagogical decision-making become tightly entangled, which limits the tutoring system’s capacity for deliberate instructional adaptation.
We propose SLOW, a theory-informed tutoring framework that supports deliberate learner-state reasoning within a transparent decision workspace. Inspired by dual-process accounts of human tutoring, SLOW explicitly separates learner-state inference from instructional action selection. The framework integrates causal evidence parsing from learner language, fuzzy cognitive diagnosis with counterfactual stability analysis, and prospective affective reasoning to anticipate how instructional choices may influence learners’ emotional trajectories. These signals are jointly considered to guide pedagogically and affectively aligned tutoring strategies.
Evaluation using hybrid human-AI judgments demonstrates significant improvements in personalization, emotional sensitivity, and clarity. Ablation studies further confirm the necessity of each module, showcasing how SLOW enables interpretable and reliable intelligent tutoring through a visualized decision-making process. This work advances the interpretability and educational validity of LLM-based adaptive instruction.

\keywords{Intelligent Tutoring Systems\and Multi-Agent Systems\and Learner Modeling\and Explainable AI in Education}
\end{abstract}
\section{Introduction}
\label{sec1}

The ``iron triangle'' among scale, personalization, and quality has long been one of the central challenges in education~\cite{ryan2021beyond}. AI has often been viewed as a possible way to ease this trade-off by enabling more scalable and adaptive forms of instructional support. More recently, the rapid advancement of LLMs has further rekindled such expectations, as their fluent natural language interaction capabilities seem to offer new possibilities for intelligent tutoring at scale. However, linguistic fluency does not equate to the capacity for educationally meaningful reasoning and instructional decision-making.

In educational technology, efforts to address this tension have traditionally been embodied in Intelligent Tutoring Systems (ITS)~\cite{murray2003overview} and adaptive learning systems (ALS)~\cite{kabudi2021ai}. These systems typically rely on explicit learner modeling and instructional strategy adjustments to achieve personalized support. As LLMs are introduced into educational contexts, however, generative models are increasingly expected to perform both learner-state diagnosis and instructional response generation within the same conversational agent~\cite{liu2024personality}. This shift blurs the boundary between learner-state diagnosis and pedagogical decision-making, creating a fundamental challenge for LLM-based tutoring: supporting explicit learner-state reasoning that can guide deliberate instructional planning~\cite{rooein2026pats}.

\begin{figure}[t]
    \centering
    \includegraphics[width=\linewidth]{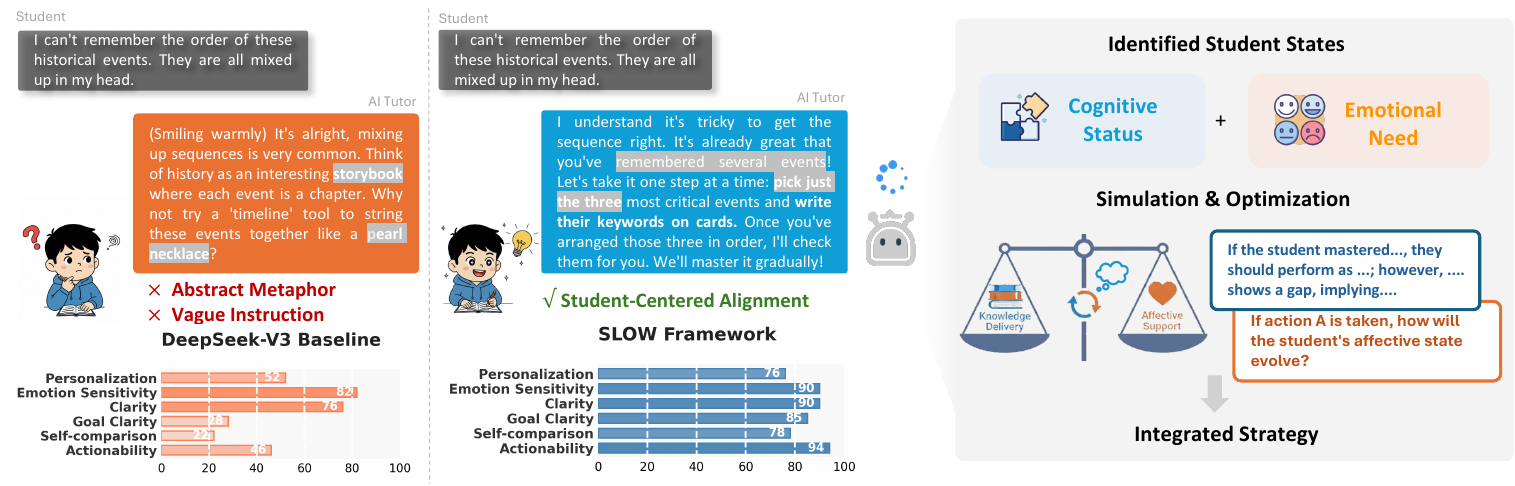}
    \caption{\textbf{An illustrative case motivating the need for deliberate learner-state reasoning.} 
    General LLMs (Left) often provides abstract Metaphors because their "fast-thinking" nature leads to the entanglement of diagnosis and decision-making. 
    Our SLOW framework (Right) introduces a transparent open workspace to decouple these processes. By performing counterfactual cognitive validation and prospective affective simulation, SLOW explicitly weighs cognitive gains against affective risks to generate guidance that is more aligned with the student's needs.}
    \label{fig:intro_case}
\end{figure}

In most existing LLM-driven tutoring systems, learner-state inference and instructional action decision-making are compressed into a single generation process. As a result, multiple signals, including cognitive diagnosis, affect perception, and pedagogical intent, are handled in a highly coupled manner, making it difficult to explicitly verify diagnostic hypotheses, compare alternative pedagogical strategies, or maintain alignment between inferred learner needs and generated instructional responses. Figure~\ref{fig:intro_case} illustrates a typical failure scenario: when a learner expresses confusion about the chronological order of historical events, generic LLM tutoring tends to provide abstract metaphors or vague encouragement, revealing a clear pedagogical misalignment between the system's response and the learner's actual cognitive difficulty.

To address these issues, existing research has mainly proceeded along two paths. One is the theory-driven paradigm, which improves the normativity of instructional strategies by translating educational or cognitive-psychological theories into model constraints, as exemplified by SocraticLM~\cite{liu2024socraticlm}, KELE~\cite{peng2025kele}, and PATS~\cite{rooein2026pats}. However, such approaches rely heavily on expert-defined rules and often struggle to generalize across domains or long-tail learning behaviors. The other is the simulation-based paradigm, which enhances system perception by constructing virtual learners, such as personality-trait-based learner modeling~\cite{liu2024personality} and dialogue trajectory simulation~\cite{jia2025diacdm,zhang2025simulating}. Yet because these simulations are largely driven by the probabilistic generation of LLMs, their diagnostic logic remains difficult to verify against educational measurement and cognitive principles, thereby limiting interpretability and pedagogical credibility. Despite their different implementation paths, both paradigms generally retain a tightly coupled generation process in which learner diagnosis and instructional action are resolved simultaneously. This differs from expert teaching practice, where teachers typically elicit and interpret evidence about learner understanding and then adjust instructional actions iteratively through formative assessment~\cite{Black01031998}. As a result, existing systems often act on insufficiently verified learner models and struggle to support timely state correction during interaction~\cite{jia2025diacdm}, which can lead to persistent strategy misalignment and instructional inertia across turns.

To address this, we propose SLOW, a ``slow thinking'' intelligent tutoring framework inspired by dual-system theory~\cite{frankish2010dual}. Rather than directly generating instructional responses, SLOW explicitly externalizes learner state reasoning into an open reasoning workspace, structurally decoupling the processes of cognitive diagnosis and instructional action selection. This workspace operates through four collaborative reasoning stages: \textit{evidence parsing}, \textit{cognitive validation}, \textit{affect prediction}, and \textit{strategy integration}, responsible for extracting causally relevant evidence, validating the stability of learning states, estimating affective evolution risks, and balancing cognitive gains against affective risks to generate instructional actions. By explicitly reasoning about learner states and simulating instructional consequences before response generation, SLOW provides a transparent, traceable, and educationally rational decision path for intelligent tutoring, thereby establishing a more interpretable and trustworthy architectural foundation for LLM-driven tutoring systems.

The main contributions of this paper are as follows:
\begin{enumerate}
    \item We reveal a structural limitation in current LLM-driven tutoring systems: the lack of explicit reasoning workspaces leads to coupling between cognitive diagnosis and instructional planning, resulting in inexplicable and error-prone instructional behaviors;
    \item We propose SLOW, a learner-centered reasoning architecture that simulates expert teachers' deliberate instructional processes through an open reasoning workspace, achieving interpretable learner diagnosis and adaptive instruction within a unified framework;
    \item Through theory-driven empirical evaluation and human-AI collaborative rating, we show that SLOW improves in instructional personalization, affective sensitivity, clarity, and operability, demonstrating how its traceable reasoning paths support educational effectiveness and trustworthy deployment.
\end{enumerate}

\section{Related Work}
\label{sec:related_work}

\subsection{Dual-System Theory and Test-Time Scaling}
The Dual-System Theory serves as a cornerstone of cognitive psychology, bifurcating human cognition into System 1, which is intuitive and automatic (``fast thinking''), and System 2, which is logical and deliberative (``slow thinking'') \cite{frankish2010dual}. As the research focus of LLMs shifts from the ``scaling laws'' of parameters to \textit{test-time scaling}~\cite{muennighoff2025s1}, which increases computational steps during inference, this theory has been revitalized in the field of artificial intelligence. Frontiers such as OpenAI o1 have demonstrated that by incorporating Chain-of-Thought (CoT)~\cite{wei2022chain} and search algorithms, models can significantly enhance their logical deduction capabilities in tasks with objective standards, such as mathematics and programming. However, existing research on test-time scaling is predominantly concentrated on domains with explicit feedback loops~\cite{ryan2021beyond}. In the context of ITS, which involve high interactive complexity, utilizing test-time computation to enhance the perception of learner states and pedagogical planning remains an under-explored frontier.

\subsection{LLM-based ITS}
Existing efforts to enhance pedagogical planning in LLM-based ITS can be broadly grouped into two paradigms. The theory-driven paradigm translates educational or cognitive theories into explicit behavioral constraints. For example, SocraticLM \cite{liu2024socraticlm} and KELE \cite{peng2025kele} use supervised fine-tuning or multi-agent strategies to strengthen guided questioning, while PATS \cite{rooein2026pats} explicitly maps learner profiles to instructional strategies. Although these methods improve the normativity of instructional behavior, they often depend heavily on expert-defined rules and struggle to generalize across domains or long-tail learner behaviors. The simulation-based paradigm improves system perception by constructing virtual learners. Prior work has used personality-based modeling to simulate diverse student responses \cite{liu2024personality}, simulated misconceptions to improve error correction, and adopted frameworks such as SimTutor~\cite{manh2025synthesizing} to preview dialogue trajectories. However, because these simulations are largely driven by LLM generation rather than explicit cognitive diagnosis model(CDM), it remains difficult to verify whether they align with psychometric principles, which limits diagnostic explainability in complex tutoring settings.

\subsection{Dialogue-based Learner Modeling}
Learner modeling underpins adaptive education and has evolved from static assessment based on controlled testing to dynamic inference from natural interaction. Early neural cognitive diagnosis models, such as NeuralCDM~\cite{wang2020neural}, established a basis for learner-state estimation, but remained limited in modeling complex knowledge relations. More recent work has sought to improve robustness and transparency. For instance, tFCM~\cite{wei2023interpretable} combines Temporal Fuzzy Cognitive Maps with the Markov Blanket principle to support causally interpretable knowledge-state transitions. With the rise of LLMs, researchers have further explored learner modeling from dialogue. Dialogue-KT~\cite{scarlatos2025exploring} uses conversational context to track knowledge mastery, but such generative approaches often lack explicit educational-measurement constraints, reducing pedagogical interpretability. DiaCDM~\cite{jia2025diacdm} further introduces diagnostic signal extraction based on the Initiation--Response--Evaluation framework. However, its design is oriented more toward offline post hoc analysis than online interaction, making it difficult to capture rapid learner-state shifts during tutoring. As a result, existing approaches still struggle to jointly support transparent reasoning and online diagnosis in open-ended dialogue.

\section{Methods}
\begin{figure}[t]
    \centering
    \includegraphics[width=\linewidth]{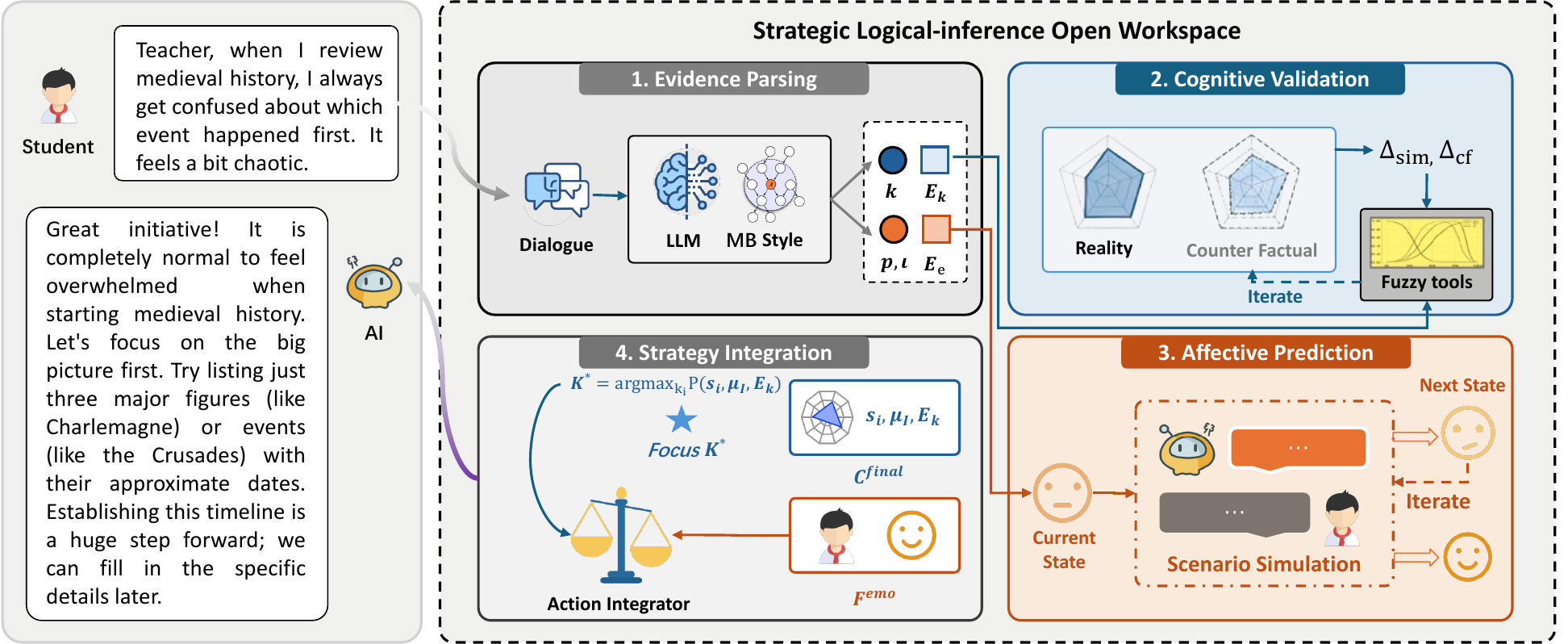}
    \caption{Overview of the Strategic Logical-inference Open Workspace (SLOW) framework. The architecture illustrates the transition from (1) Evidence Parsing, where dialogue is deconstructed into cognitive and affective primitives, to (2) Cognitive Validation and (3) Affective Prediction, which utilize counterfactual simulation and prospective simulation to refine the internal state. Finally, (4) Strategy Integration balances these signals to execute a calibrated tutoring action.}
    \label{fig:methods}
\end{figure}
The SLOW (Strategic Logical-inference Open Workspace) framework (Figure \ref{fig:methods}) facilitates pedagogical reasoning by constructing an explicit inference space within a generative architecture. This design is grounded in the principles of formative assessment, which, as noted by Sadler~\cite{sadler1989formative}, necessitates a clear understanding of a learner's current state before taking pedagogical action to close learning gaps. To operationalize this theory, SLOW simulates the internal deliberative mechanism of an expert teacher who performs structured analysis and anticipates potential consequences within a mental workspace rather than responding impulsively. Accordingly, the framework deconstructs this psychological process into four synergistic stages:\textbf{ parsing evidence, cognitive validation, affective prediction, and strategy integration}.

\subsection{Parsing Evidence}

To mitigate the inherent linguistic noise and redundancy present in open-ended learner discourse, the SLOW framework initiates its reasoning process with a structured evidence extraction stage. Grounded in the \textbf{Markov Blanket (MB)}~\cite{fu2010markov} principle, we identify a minimal sufficient feature set for each Knowledge Component (KC) $k$. This ensures that once these diagnostic features are determined, the mastery state of $k$ remains conditionally independent of peripheral dialogue elements.

The system decomposes the raw input $u^{\text{orig}}$ into two distinct diagnostic streams: cognitive primitives $(k, E_k)$ and affective triplets $(p, \iota, E_e)$. Specifically, each identified KC $k$ and its associated evidence span $E_k$ is transformed into a dense \textbf{diagnostic encoding} $\mathbf{z}_k$:
\begin{equation}
\mathbf{z}_k = \Phi_{\text{MB}}(k, E_k; \mathcal{S}_{\text{default}}^{\text{MB}})
\end{equation}
where $\mathcal{S}_{\text{default}}^{\text{MB}}$ represents a specialized set of MB-style feature templates designed to capture mastery-relevant indicators directly from natural language. Unlike traditional methods that rely on explicit behavioral logs, the mapping function $\Phi_{\text{MB}}$ derives latent proxies of mastery by analyzing the causal relevance of the learner's expressed reasoning within the current interaction context.

When the input contains emotional cues, the system extracts the polarity $p$, intensity $\iota$, and supporting text $E_e$ to form an affective vector $\mathbf{e}$. This vector serves as the initial affective baseline for the subsequent simulation of emotional trajectories. The comprehensive output of this stage is the label set $\mathcal{T}(u^{\text{orig}})$, which consolidates the collection of cognitive features $Z = \{\mathbf{z}_k\}$ and the initial affective state $\mathbf{e}$ for downstream deliberative analysis.

\subsection{Cognitive Validation}

Upon obtaining the structural features $\mathbf{z}_k$, the system initiates an analysis of the learner's cognitive state. Recognizing that knowledge acquisition is a non-discrete evolutionary process, we utilize a Fuzzy Cognitive Discriminator to represent the mastery level $\mu(C_k)$~\cite{wei2023interpretable}. The state is quantified as a continuous membership distribution across four hierarchical levels:
\begin{equation}
\mu(C_k) = (\mu_{\text{Un}}, \mu_{\text{InK}}, \mu_{\text{K}}, \mu_{\text{L}})
\end{equation}
where the states correspond to Unknown (Un), Insufficiently Known (InK), Known (K), and Learned (L), respectively.

To evaluate the stability of this diagnosis, the system executes a Counterfactual Simulation within the deliberative workspace. This process assesses the robustness of the observed state by constructing a counterfactual hypothesis (e.g., ``Assume the learner already possesses state K'') and deducing the corresponding typical features $F_k^{\text{sim}}$.

The system evaluates the alignment between the empirical reality $F_k^{\text{orig}}$ and the simulated profile $F_k^{\text{sim}}$ by calculating two contrastive signals:
\begin{equation}
\Delta_{\text{sim}} = \text{Diff}(F_k^{\text{orig}}, F_k^{\text{sim}})
\end{equation}
where $\Delta_{\text{sim}}$ identifies the directional mismatch signal. These signals, along with the counterfactual effort $\Delta_{\text{cf}}$ required to shift between states, are then processed by \textbf{Fuzzy tools}~\cite{wei2023interpretable} to iteratively refine the diagnostic score. This validation loop ensures that the final output reaches a stable cognitive context $C^{\text{final}}$, thereby enhancing the transparency and interpretability of the modeling process.

\subsection{Affective Prediction}

Simultaneously, the system activates prospective affective simulation to anticipate the impact of potential pedagogical interventions. As illustrated in Figure \ref{fig:methods} Block 3, the simulator utilizes the emotional triplets extracted during the parsing stage to initialize the \textbf{Current State} $e_{\text{before}}$, representing the learner's baseline polarity and intensity. 

To evaluate the emotional trajectory, the system conducts a forward rollout by simulating a pool of candidate tutor responses $\{r^{(1)}, \dots, r^{(M)}\}$ to predict the learner's emotional \textbf{Next State} $e_{\text{after}}^{(m)}$ for each candidate $r^{(m)}$. A transition score $\Delta^{(m)}$ is calculated to evaluate the affective shift between the current state $e_{\text{before}}$ and the predicted state $e_{\text{after}}^{(m)}$:
\begin{equation}
\Delta^{(m)} = \text{Score}(e_{\text{before}}, e_{\text{after}}^{(m)})
\end{equation}
where $m$ denotes the index of the response draft being evaluated. The optimal simulation result is then synthesized into the final \textbf{Affective Prediction} signal, encoded as a control vector $F^{\text{emo}} = (\text{emo}_{\text{cur}}, \text{int}_{\text{cur}}, \text{tgt}_{\text{cur}})$. In this formulation, $\text{emo}_{\text{cur}}$ and $\text{int}_{\text{cur}}$ specify the target polarity and intensity, while $\text{tgt}_{\text{cur}}$ identifies a prescriptive control target (e.g., ``encourage'' or ``stabilize''). This prospective mechanism ensures that the instructional path is motivationally supportive by mitigating the risk of triggering learner frustration before the response is actually delivered.

\subsection{Strategy Integration}

In the final stage, the system integrates the simulated cognitive and affective signals to facilitate a multi-criteria decision-making process. As illustrated by the Figure \ref{fig:methods}, Block 4, the framework identifies the optimal pedagogical focus $k^*$ by ranking candidates according to their priority scores:
\begin{equation}
k^* = \arg\max_{k_i} \text{Priority}(s_i, \mu_i, E_{k_i})
\end{equation}
This decision logic performs a critical trade-off among three primary dimensions for each candidate knowledge component $k_i$. Specifically, the integrator evaluates the mastery severity $s_i$ along the pedagogical hierarchy (Un → InK → K → L), the diagnostic confidence $\mu_i$ derived from the membership stability and counterfactual analysis, and the richness of supporting evidence $E_{k_i}$ extracted from the dialogue.

The associated state $s^*$ of the selected focus $k^*$ subsequently determines the instructional stance, ranging from foundational scaffolding for state Un to transfer-oriented extension for state L. To generate the final tutoring response $y^{\text{resp}}$, the system couples the finalized cognitive diagnosis $C^{\text{final}}$ with the affective control vector $F^{\text{emo}}$. By balancing instructional necessity with emotional stability on a conceptual scale, the framework ensures that the resulting tutoring action is both cognitively precise and motivationally supportive.

\section{Experimental Design}
This section details the experimental configuration, covering the construction of the evaluation dataset, the selection and configuration of baseline models, and the evaluation mechanism.

\subsection{Dataset}
The evaluation dataset is designed to capture representative tutoring challenges in authentic instructional contexts. It combines authentic student--teacher interaction data with model-augmented samples. The source data were drawn from real interaction corpora, reflecting typical cognitive hurdles and affective expressions, such as confusion over circuit current directions, fragmented knowledge acquisition, and learner doubt with affective undertones. To increase diversity, we used a large language model to generate 200 additional samples conditioned on these authentic inquiries, while preserving the original knowledge topics and difficulty levels and varying surface formulations and affective tones. All candidate instances were then manually reviewed by educational experts to remove ambiguous or pedagogically unrealistic cases, yielding a final set of 100 expert-validated high-quality instances.

The final dataset spans K1--K12 and covers seven disciplines: Biology (20), Physics (20), Mathematics (20), History (14), Geography (12), Chemistry (10), and English (4). It also balances five scenario types---Affective Support (32), Personalized Support (26), Strategic Scaffolding (22), Direct Q\&A (12), and Error Correction (8)---as well as three emotion categories: Positive (36), Neutral (32), and Negative (32). Together, these properties ensure pedagogical realism, cross-disciplinary coverage, and controlled variation in learner states.

\subsection{Baselines}
To verify the generalizability of SLOW across different foundation models, we selected representative models from three major families: GPT-4o and GPT-4o-mini from OpenAI, Gemini-1.5-Pro and Gemini-1.5-Flash from Google, and DeepSeek-V3 alongside the reasoning-enhanced DeepSeek-R1. 

To ensure a rigorous and fair comparison, all baselines are configured with strong prompting and follow a two-stage pipeline where the model first explicitly diagnoses the learner's cognitive and affective states before generating a final response. Critically, the baseline prompts explicitly incorporate the full set of evaluation rubrics. This setup ensures that baselines are fully aware of the scoring preferences, thereby confirming that any performance gains from SLOW stem from its internal open reasoning workspace and simulation mechanisms rather than mere prompt engineering or information disparity. Baseline prompt has been open-sourced at \url{https://github.com/PhilrainV/SLOW}.

\subsection{Metrics}

Tutoring responses are evaluated using a principle-driven framework grounded in cognitive load theory \cite{sweller2011cognitive} and formative feedback principles \cite{sadler1989formative}. Rather than relying on reference-based matching, which is ill-suited to open-ended tutoring where effective strategies are inherently non-unique, our framework assesses response quality directly with respect to pedagogical appropriateness and cognitive efficiency.

Each tutoring response is evaluated along seven response-level dimensions. These dimensions operationalize core aspects of effective tutoring, including diagnostic appropriateness, controlled cognitive load, and guidance toward concrete next steps. The framework explicitly penalizes excessive verbosity, redundant explanations, and multiple parallel solution paths that increase cognitive and decision-making load, while favoring concise, focused responses that align with the learner’s expressed understanding and instructional needs and propose a minimal actionable step. Table~\ref{table:rubrics} summarizes the complete evaluation rubric.

\begin{table}[tb]
\centering
\footnotesize
\caption{Evaluation rubric for personalized tutoring quality.}
\label{table:rubrics}

\renewcommand{\arraystretch}{1}

\begin{tabular*}{\textwidth}{@{\extracolsep{\fill}} m{2.8cm} p{9cm}}
\toprule

\textbf{Dimension} & \textbf{Definition} \\ 
\midrule

\textcolor{DimBlue}{Clarity} 
& The response is easy to understand, well-structured, and unambiguous. Excessive verbosity, redundant explanations, or unnecessary enumeration are penalized due to increased cognitive load. \\ 
\cmidrule(l){2-2}

\textcolor{DimOrange}{Goal Clarity} 
& The response makes its instructional intent explicit, enabling the learner to clearly understand the immediate learning objective for the current turn. \\ 
\cmidrule(l){2-2}

\textcolor{DimGreen}{Emotion Sensitivity} 
& The response appropriately attends to emotional cues expressed in the learner’s utterance, providing reassurance, encouragement, or neutral guidance when appropriate, without exaggerated or unnecessary affective language. \\ 
\cmidrule(l){2-2}

\textcolor{DimPurple}{Self-comparison} 
& The response frames feedback in terms of the learner’s own progress and remaining gaps, emphasizing personal improvement rather than peer comparison or competitive evaluation. \\ 
\cmidrule(l){2-2}

\textcolor{DimRed}{Personalization} 
& The response is tailored to the learner’s expressed difficulty or apparent level of understanding, avoiding generic, template-based, or broadly applicable explanations. \\ 
\cmidrule(l){2-2}

\textcolor{DimTeal}{Actionability} 
& The response provides a specific, minimal, and immediately executable next step. Responses that present multiple parallel options are penalized for increasing decision-making and cognitive load. \\ 
\midrule

\textbf{Overall Score} 
& A holistic judgment of tutoring quality, reflecting instructional usefulness, emotional appropriateness, and effective management of cognitive load. \\

\bottomrule
\end{tabular*}
\end{table}

The evaluation protocol follows a single-blind procedure where human evaluators remain unaware of the model identity behind each response. Each response is independently rated by two experts and an automated judge (GPT-5 equivalent model) using the 0--100 scale rubrics. The final score is derived from an equally weighted average of human and LLM ratings.

\section{Results}

Our evaluation analyzes the performance of the SLOW framework from four distinct perspectives: 
(i) a \textbf{multi-model comparison} evaluating pedagogical gains over baseline models across diverse dimensions; 
(ii) an \textbf{ablation study} to determine the contribution of individual architectural components; 
(iii) a \textbf{computational efficiency analysis} assessing the trade-off between reasoning overhead and instructional quality; 
and (iv) an \textbf{interpretability analysis} demonstrating the transparent reasoning process within the workspace.

\begin{table}[tb]
    \centering
    \caption{Comparative analysis of performance gains: score differences between SLOW and the baseline model across tutoring dimensions.}
    \label{tab:model_comparison}
    \begin{tabular*}{\textwidth}{@{\extracolsep{\fill}}lccccccc}
        \toprule
        \textbf{Model} & 
        \textbf{\textcolor{DimBlue}{$\Delta$Clar.}} & 
        \textbf{\textcolor{DimOrange}{$\Delta$Goal.}} & 
        \textbf{\textcolor{DimGreen}{$\Delta$Emo.}} & 
        \textbf{\textcolor{DimPurple}{$\Delta$SelfComp.}} & 
        \textbf{\textcolor{DimRed}{$\Delta$Pers.}} & 
        \textbf{\textcolor{DimTeal}{$\Delta$Act.}} &
        \textbf{$\Delta$Overall} \\ 
        \midrule
        deepseek-r1      & \plus{41.4} & \plus{42.2} & \plus{20.4} & \plus{47.2} & \plus{19.4} & \plus{62.4} & \plus{38.0} \\
        deepseek-v3      & \minus{6.80} & \plus{32.2} & \plus{5.40} & \plus{25.4} & \plus{7.20} & \plus{17.0} & \plus{10.6} \\
        gemini-2.5-flash & \plus{31.4} & \plus{48.6} & \plus{28.2} & \plus{39.2} & \plus{19.0} & \plus{36.8} & \plus{29.8} \\
        gemini-2.5-pro   & \plus{35.6} & \plus{37.4} & \plus{18.8} & \plus{25.2} & \plus{17.8} & \plus{40.6} & \plus{30.0} \\
        gemini-3-pro     & \plus{28.8} & \plus{22.0} & \plus{14.2} & \plus{25.4} & \plus{11.0} & \plus{21.8} & \plus{20.4} \\
        gpt-4.1          & \plus{42.2} & \plus{36.2} & \plus{15.2} & \plus{32.6} & \plus{12.6} & \plus{55.0} & \plus{20.9} \\
        gpt-4.1-mini     & \plus{22.2} & \plus{33.0} & \plus{20.0} & \plus{31.2} & \plus{9.20} & \plus{25.8} & \plus{14.2} \\
        gpt-4o           & \plus{29.4} & \plus{33.8} & \plus{14.6} & \plus{28.2} & \plus{10.0} & \plus{39.4} & \plus{20.8} \\
        gpt-4o-mini      & \plus{14.8} & \plus{20.0} & \plus{12.8} & \plus{23.0} & \plus{4.00} & \plus{16.2} & \plus{8.6} \\
        \bottomrule
    \end{tabular*}
\end{table}

\subsection{Model Comparison}

Table~\ref{tab:model_comparison} summarizes the performance gains of SLOW over the prompt-based baseline across six tutoring dimensions. Across all model families, SLOW consistently improves response quality on nearly all dimensions, with similar improvement patterns observed for both large and compact models, suggesting that the gains are attributable to the framework’s structural design rather than being solely driven by model capacity. 
To assess the robustness of these results, we conducted Wilcoxon signed-rank tests for each of the nine backbone models over the evaluation instances. SLOW outperformed the baseline significantly for all nine models, with $p < 0.001$ for seven models and $p < 0.01$ for the remaining two. Furthermore, Cliff's $\delta$ values ranged from 0.42 to 0.59, indicating medium-to-large effect sizes.
Notable improvements are observed in clarity and goal clarity, particularly for DeepSeek-R1 and GPT-4.1, suggesting that SLOW improves instructional focus and response clarity. A small degradation in clarity is observed for DeepSeek-V3 \cite{liu2024deepseek}, possibly reflecting its tendency toward more verbose intermediate reasoning. In contrast, large gains in actionability (e.g., +62.4 for DeepSeek-R1\cite{guo2025deepseek}) highlight SLOW’s effectiveness in translating diagnostic insights into concrete next-step guidance.

An ablation study compares the full SLOW framework with variants removing Cognitive Validation or Affective Prediction. As shown in Table~\ref{tab:ablation}, the full system performs best across all models. Removing Cognitive Validation generally causes substantial degradation, especially for DeepSeek-R1, while removing Affective Prediction also leads to clear performance losses across models. These results suggest that cognitive validation and affective prediction provide complementary benefits for effective tutoring.

\begin{table}[tb]
    \centering
    \caption{\textbf{Overall Score} in the ablation study. ``w/o'' denotes ``without''.}
    \label{tab:ablation}
    \begin{tabular}{lccc}
        \toprule
        Ablation Setting & GPT-4.1 & Gemini-3-Pro & DeepSeek-R1 \\
        \midrule
        Baseline        & 59.6 & 62.0 & 50.2\\
        SLOW w/o Cognitive Validation     & 73.0 & 78.6 & 72.8\\
        SLOW w/o Affective Prediction    & 77.2 & 77.8 & 66.0\\
        SLOW (Full)     & \textbf{83.0} & \textbf{88.6} & \textbf{89.6}\\
        \bottomrule
    \end{tabular}
\end{table}

Because these comparisons rely on rubric-based evaluation, we further examined the reliability of the scoring framework. Specifically, We computed Cronbach's $\alpha$ across rubric dimensions as an index of internal consistency for the scoring framework. For the two human experts, the ratings showed good internal consistency ($\alpha = 0.84$), while the hybrid human--AI evaluation (including the LLM rater) also maintained good internal consistency ($\alpha = 0.81$). Inter-rater agreement between the two human experts was measured by ICC(2,1), yielding a value of $0.78$. In addition, the rank-order alignment between human and LLM judgments reached a Spearman's $\rho = 0.72$. These results indicate that the scoring framework is sufficiently reliable.

\subsection{Computational Efficiency and Cost}
To evaluate the trade-off between pedagogical gain and computational overhead, we conducted a cost analysis using GPT-4o-mini as the backbone model with deterministic decoding ($\text{temperature}=0$). Relative to the standard Baseline (1.0$\times$cost, 65.4 overall score) and the EduPlanner-style framework \cite{zhang2025eduplanner} (3.6$\times$cost, 73.2 overall score), SLOW incurs a computational cost of 6.4$\times$ and achieves an overall score of 79.6.

Notably, SLOW does not rely on a fixed-length reasoning chain; additional iterations are triggered only when diagnostic inconsistencies are detected, yielding a median of 6 and an 80th percentile of 7 API calls per instance. To test whether these gains are merely due to increased inference-time compute, we compared SLOW against a compute-matched 7-step refinement control (Refine-7). Refine-7 uses seven sequential calls (draft, critique, revision, critique, revision, critique, and final revision) with the same backbone model, decoding setting, and evaluation rubric, resulting in a similar cost of 6.2$\times$. Unlike SLOW, Refine-7 does not maintain explicit learner-state representations or implement explicit cognitive--affective decomposition, but instead relies on unconstrained multi-turn refinement. Despite the closely matched budget (within 5\% of SLOW's token budget), Refine-7 achieved an overall score of only 71.2, substantially lower than SLOW's 79.6. These results suggest that SLOW's gains are not explained by additional compute alone, but by its structured pedagogical architecture.

\subsection{Interpretability Analysis}

\begin{figure}[tb]
    \centering
    \includegraphics[width=\linewidth]{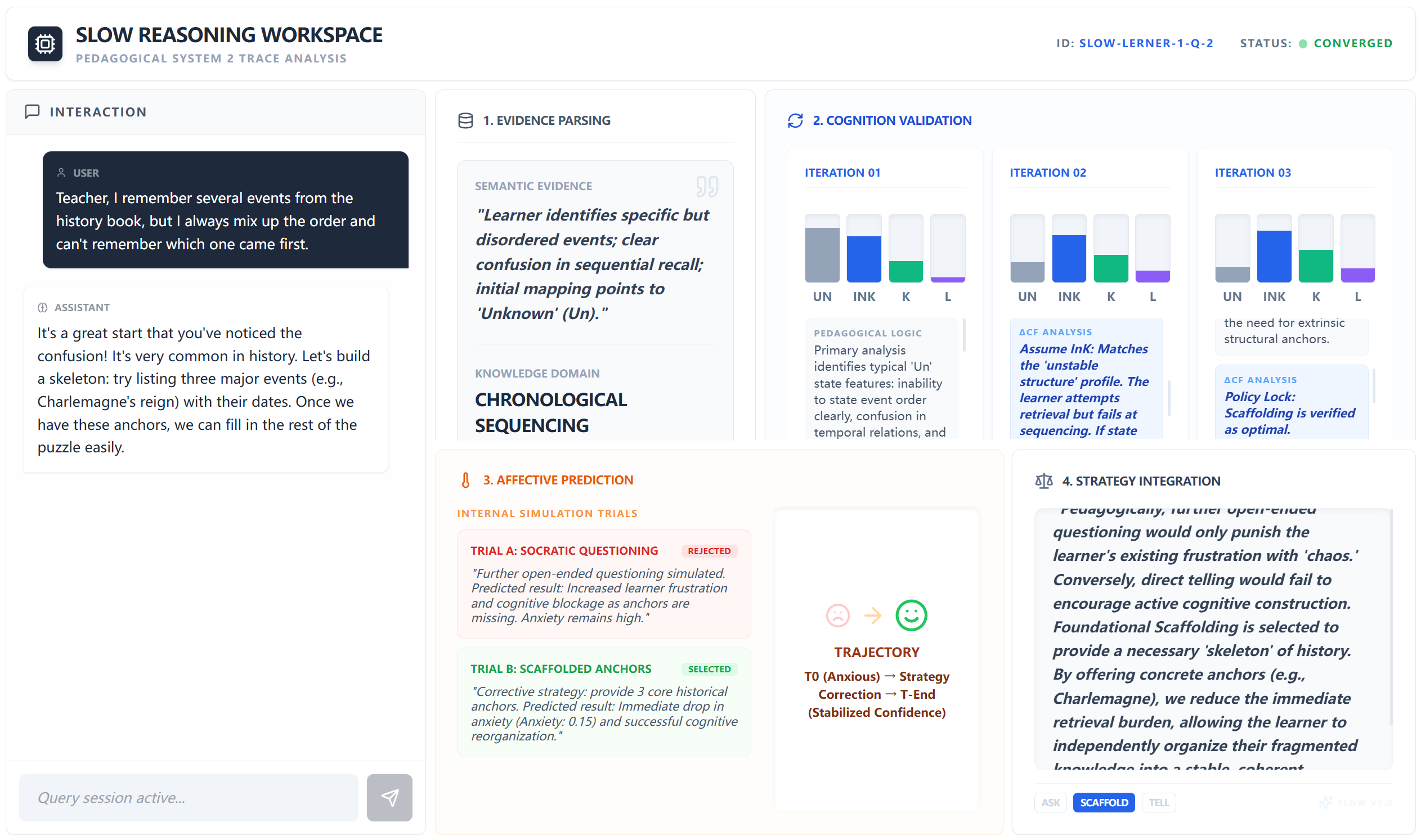}
    \caption{A case of SLOW Reasoning Workspace demonstrating interpretability.}
    \label{fig:case}
\end{figure}

To demonstrate the interpretability of SLOW beyond surface-level explanations, we developed an interactive Reasoning Workspace (Figure~\ref{fig:case}) that externalizes the system’s internal deliberation as a human-readable System~2 pedagogical trace. Rather than providing post-hoc rationales for a generated response, the workspace exposes the intermediate diagnostic assumptions, counterfactual evaluations, and strategy trade-offs that shape instructional decisions.
As illustrated in the historical sequencing case, the workspace supports interpretability at three complementary levels that are meaningful:

\begin{itemize}
    \item \textbf{Diagnostic Traceability}. When the learner reports difficulty recalling event order, the workspace reveals how the system parses this utterance into a specific cognitive hypothesis---namely, a deficit in chronological structuring rather than factual recall. Through iterative cognitive validation, the learner’s mastery profile is refined from an initial \emph{Unknown} (Un) state to \emph{Insufficiently Known} (InK), with explicit evidence indicating fragmented knowledge and missing structural anchors. This allows human observers to inspect the final diagnosis and the evidential path leading to it.

    \item \textbf{Risk-Aware Strategy Selection}. The workspace makes instructional deliberation explicit by displaying candidate strategies that were considered and rejected. In this case, a Socratic questioning approach is discarded because prospective simulation predicts heightened anxiety and cognitive blockage given the learner’s current state. By exposing these alternatives, the system clarifies why certain pedagogically plausible actions are intentionally avoided, supporting informed auditing of instructional risk management.

    \item \textbf{Calibrated Feedback Rationale}. Finally, the workspace provides a plain-language justification for the selected action. It explains how supplying minimal chronological anchors reduces immediate retrieval load and enables the learner to reorganize fragmented knowledge independently. This rationale connects diagnostic conclusions to concrete instructional choices in a manner accessible to both teachers and learners.
\end{itemize}

By decomposing tutoring behavior into observable diagnostic, affective, and strategic layers, the SLOW Reasoning Workspace moves beyond black-box automation. It functions as an auditable pedagogical interface that allows stakeholders to understand, evaluate, and trust how instructional decisions are formed, thereby supporting both educational validity and responsible deployment of LLM-based tutors.

\section{Discussion \& Conclusion}

This paper proposes the SLOW framework, which introduces an open reasoning workspace to enable systematic consideration of learners’ cognitive and affective factors prior to instructional response generation. Based on this design, we construct a structured reasoning process consisting of evidence parsing, cognitive validation, affective reasoning, and strategy integration. Empirical evaluation demonstrates that SLOW improves instructional specificity, actionability, and pedagogical coherence.

Alongside these improvements, the proposed design also introduces corresponding costs. By incorporating explicit reasoning and simulation during interaction, SLOW may result in longer response latency and could inadvertently amplify internal model biases toward diverse learner profiles. Consequently, future work should explore memory summarization and state-caching mechanisms to reduce redundant reasoning costs, alongside dedicated bias detection and mitigation protocols to enhance pedagogical fairness. Whether this trade-off between reasoning effort and instructional quality can be stably translated into verifiable teaching effectiveness in real educational settings remains an open question for further empirical investigation. Beyond directly supporting learners, the transparent reasoning process exposed by SLOW may also serve teachers by illustrating how instructional decisions can be formed through systematic consideration of cognition and affect. In addition, the framework can function as a reasoning-aware data synthesis mechanism, providing more  structured reasoning data for the training of next-generation educational language models.

\noindent\paragraph{\textbf{Acknowledgements.}}
This work was supported by the National Natural Science Foundation of China (Grant No. 62477012), the Natural Science Foundation of Shanghai, China (Grant No. 23ZR1418500), the AI for Science Program of the Shanghai Municipal Commission of Economy and Informatization, China (Grant No. 2025-GZL-RGZN-BTBX-01014), Major Program of Philosophy and Social Sciences Research of the Ministry of Education (Grant No. 2025JZDZ054).

\bibliographystyle{splncs04}
\bibliography{slow}

\end{document}